\documentclass[letterpaper]{article} 
\usepackage{aaai23}  
\usepackage{times}  
\usepackage{helvet}  
\usepackage{courier}  
\usepackage[hyphens]{url}  
\usepackage{graphicx} 
\usepackage{natbib}  
\usepackage{caption} 
\usepackage{algorithm}
\usepackage{algorithmic}
\usepackage{amsmath}
\usepackage{tikz,siunitx}
\usepackage{newfloat}
\usepackage{listings}
\DeclareCaptionStyle{ruled}{labelfont=normalfont,labelsep=colon,strut=off} 
\floatstyle{ruled}
\newfloat{listing}{tb}{lst}{}
\floatname{listing}{Listing}
\title{Exploring the Effectiveness of Mask-Guided Feature Modulation as a Mechanism for Localized Style Editing of Real Images (Student Abstract)}
\author {
    Snehal Singh Tomar,
    Maitreya Suin, 
    A.N. Rajagopalan 
}
\affiliations {
    Indian Institute of Technology Madras\\
    snehal@smail.iitm.ac.in, maitreyasuin21@gmail.com, raju@ee.iitm.ac.in
}
\usepackage{bibentry}

\begin{document}

\maketitle

\begin{abstract}
The success of Deep Generative Models at high-resolution image generation has led to their extensive utilization for style editing of real images. Most existing methods work on the principle of inverting real images onto their latent space, followed by determining controllable directions. Both inversion of real images and determination of controllable latent directions are computationally expensive operations. Moreover, the determination of controllable latent directions requires additional human supervision. This work aims to explore the efficacy of mask-guided feature modulation in the latent space of a Deep Generative Model as a solution to these bottlenecks. To this end, we present the SemanticStyle Autoencoder (SSAE), a deep Generative Autoencoder model that leverages semantic mask-guided latent space manipulation for highly localized photorealistic style editing of real images. We present qualitative and quantitative results for the same and their analysis. This work shall serve as a guiding primer for future work.
\end{abstract}

\section{Introduction}
The Swapping Autoencoder (SAE) \cite{park2020swapping} is a Deep Generative AE Model which learns separate latent representations for the structure ($S_{s}$) and style ($S_{t}$) information present in images. We exploit the Modulated Convolutional layer of the SAE, where the global style vector is spatially broadcasted and multiplied with the structure tensor $S_t$ for modulating the style of the generated image. We claim that, instead of repeating the same style vector for all pixel locations, if we can first localize a particular Region of Interest (ROI) and then manipulate the style content by adding noise in the latent space for only that region, it will result in a locally edited image. The resultant image often depicts coarse style edits on a ROI. To refine this output, we employ a refinement block that  acts on the coarsely edited ROI  while being guided by a discriminator to generate realistic styles. Our  framework does not require any ground truth for the locally edited images and is only weakly supervised by GT segmentation masks. Since our method does not require the identification of controllable latent directions post training, we eliminate the additional human supervision required by SOTA methods to achieve similar results.

\section{Method}
\begin{figure}
    \centering
    \includegraphics[width = 0.45\textwidth]{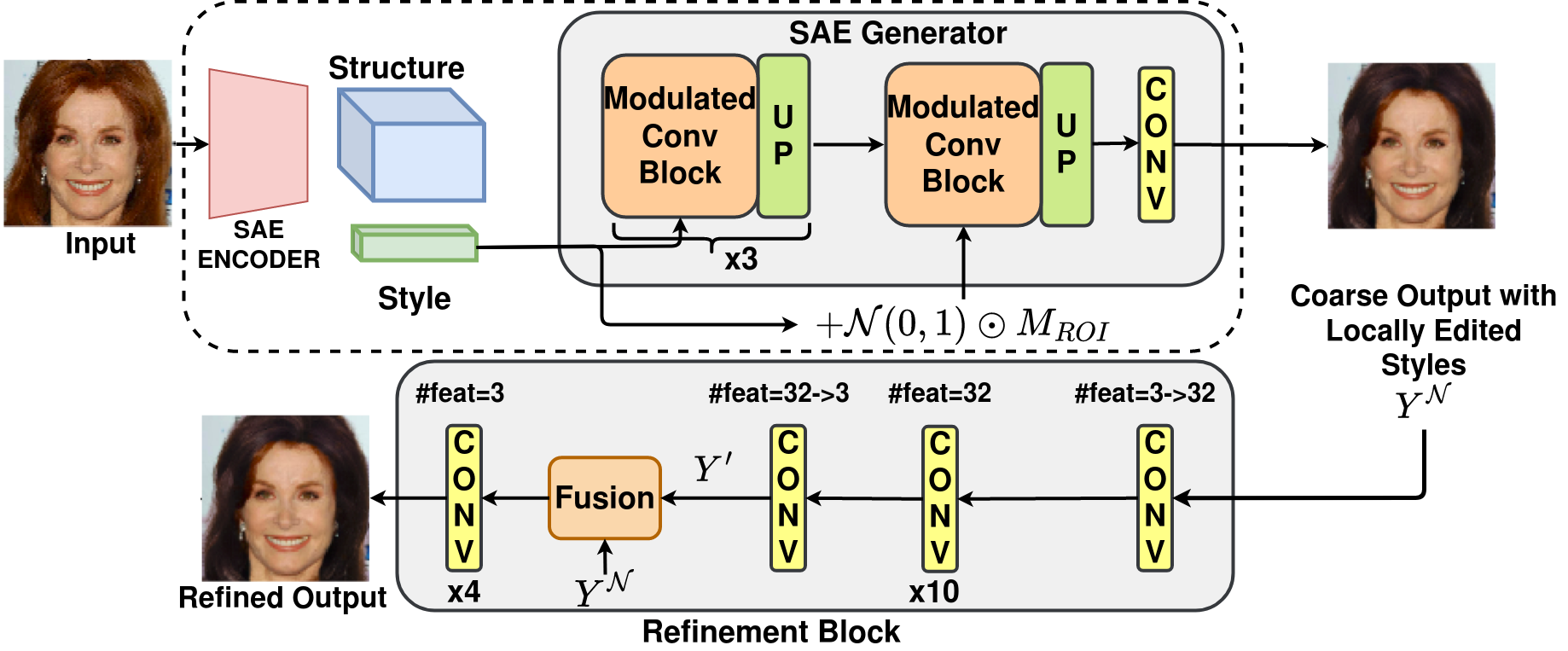}
    \caption{An illustration of the SMN (enclosed in \tikz[baseline]{\draw[densely dashed] (0,.5ex)--++(.5,0) ;}) and the Refinement Block. $M_{ROI}$ represents the SMPN's output.}
    \label{fig:overview}
\end{figure}
\begin{figure}
    \centering
    \includegraphics[width = 0.45\textwidth]{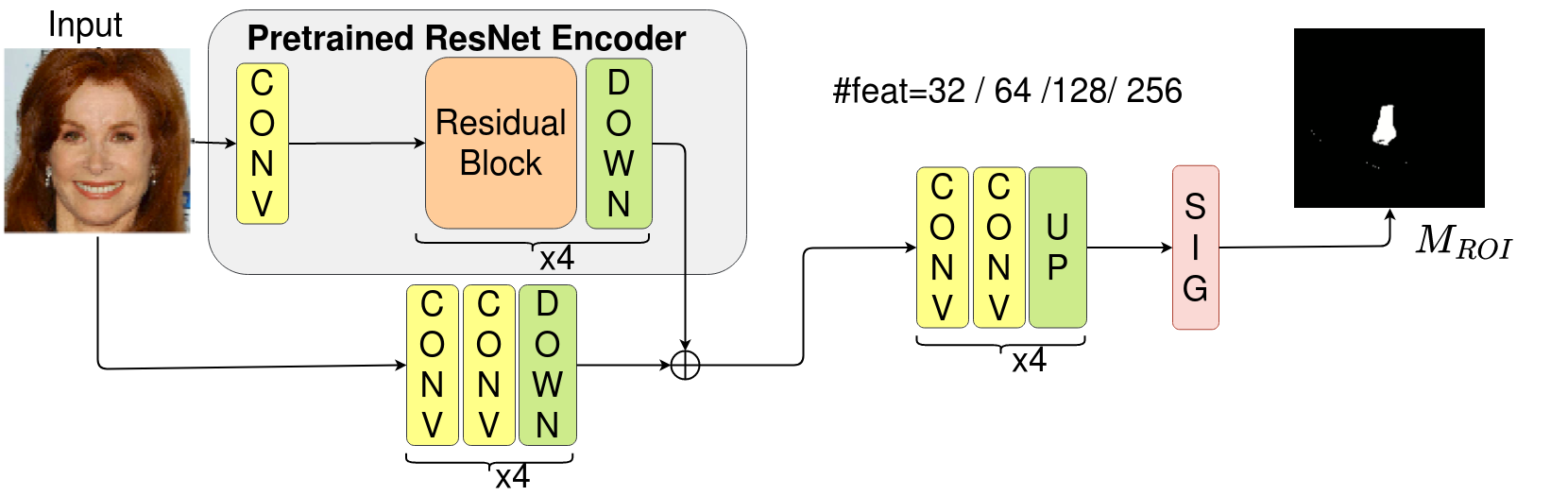}
    \caption{An overview of the Semantic Mask Prediction Network. Five networks with the same structure are used for predicting five different ROIs.}
    \label{fig:smpn}
    \vspace{-5mm}
\end{figure}

Fig. \ref{fig:overview} depicts our editing pipeline which consists of the Style Manipulation Network (SMN) and the Refinement Block. The SMN makes use of masks generated by the Semantic Mask Prediction Network (SMPN). Given an image, the SMPN first generates segmentation maps of different ROIs. Guided by these segmentation maps, the SMN performs local style modification in the input image.  
\paragraph{Semantic Mask Prediction Network (SMPN)}
 We build a semantic binary mask prediction network to localize individual regions of interest, namely: hair, skin, nose, eyes, and (lips + mouth). An overview of our SMPN is shown in Fig. \ref{fig:smpn}. We employ a pretrained ResNet18-based framework for the encoder with four levels. Further, we introduce a few parallel convolutional blocks at every level. In the decoder, we use two convolutional layers at each level, followed by an upsampling operation. In the end, we use a sigmoid operation to produce a spatial mask and train using binary cross-entropy loss with the ground-truth binary mask. We use five SMPNs with the same structure for five ROIs. 
 
\paragraph{Style Manipulation Network (SMN)}
We use a pre-trained SAE \cite{park2020swapping} to build our SMN. Given an input images of size $H \times W \times 3$, the encoder generates a structure tensor ($S_{s}$), having dimensions $H/16 \times W/16  \times 8$) and texture vector ($S_{t}$), having dimensions $1 \times 1 \times 2048$. The generator starts with $S_{s}$ and gradually upsamples and refines it to finally reconstruct the input image. The key building block of the SAE generator (decoder) is the Modulated Convolutional layer. We argue that, as the same style vector ($S_{t}$) is repeated (to form $S^{spat}_{t}$) and applied to all the pixel locations, it results in uniform style changes in the resultant image. Thus, if we vary the style vector in the spatial domain, we will be able to manipulate the local style of any particular region. Driven by this, we deploy a semantic-mask-guided style vector modulation technique in our work. Given $S^{spat}_{t}$ and a mask $M_{ROI}$ indicating a particular ROI, we perform the following operation
\begin{equation}
	S'^{spat}_{t} = S^{spat}_{t} + \mathcal{N}(0,1) \odot M_{ROI}
\end{equation}
where $\mathcal{N}(0,1)$ denotes the standard Gaussian Noise. Thus, we add ROI-specific noise to the latent style/texture vector $S'^{spat}_{t}$, which changes the style content only for the corresponding ROI. As indicated in Fig. \ref{fig:overview}, we empirically selected the last but one decoder layer to add noise and found the changes to be realistic.


\subsubsection{Refinement Block (RB)}
We refine our results using the RB. We use skip connection-based convolutional layers for the RB that operates at the full spatial resolution to preserve the finer pixel details. Let the output of the SAE after adding spatial-noise to $S_t$ be $Y^{\mathcal{N}}_{SAE}$. We apply a few convolutional operations on $Y^{\mathcal{N}}_{SAE}$ to generate $Y'_{SAE}$. Next, we fuse the refined output with the original output $Y^{\mathcal{N}}_{SAE}$ as follows
\begin{equation}
Y''_{SAE} = Y'_{SAE} \odot M_{ROI} + Y^{\mathcal{N}}_{SAE} \odot (1-M_{ROI})
\end{equation}
We further pass $Y''_{SAE}$ through a few convolutional layers to generate the final refined output $Y_{ref}$. In our work, we use five separate refinement blocks for five ROIs.
\paragraph{Objective Functions}
The Semantic Mask Prediction Network (SMPN) is trained using standard binary cross-entropy loss
\begin{equation}
\mathcal{L}_{SMPN} = \sum_{ROI} (M_{ROI},M^{GT}_{ROI})
\end{equation}
The SAE is pretrained with standard following the standard training procedure described in \cite{park2020swapping}
\begin{equation}
\begin{split}
      \mathcal{L}(Y_{SAE}, Y) = \mathcal{L}_{\text{rec}}(Y_{SAE}, Y) & + 0.5L_{\text{GAN, rec}}(Y_{SAE}, Y)\\ +  0.5L_{\text{GAN, swap}}(Y_{SAE}, Y) &  + 0.5L_{\text{CooccurGAN}}(Y_{SAE}, Y)
\end{split}
\label{eq: sae_loss}
\end{equation}
Here, $Y_{SAE}$ refers to the reconstruction obtained from the decoder for the latent space $\{S_{s}, S_{t}\}$ without any noise, $Y$ refers to the ground-truth (input) image. Other notations in Eq. \ref{eq: sae_loss} have been used exactly as in \cite{park2020swapping}. We train our refinement blocks to reconstruct the region corresponding to $1-M_{ROI}$ same as $Y^{\mathcal{N}}_{SAE}$. The style manipulated $M_{ROI}$ region is trained with standard discriminator loss. The overall loss function can be expressed as 
\begin{equation}
\begin{split}
      \mathcal{L}(Y_{ref}, Y_{SAE}) = \sum_{ROI} (\mathcal{L}_{\text{rec}}(Y_{ref} \odot (1-M_{ROI})), \\ Y_{SAE} \odot (1-M_{ROI})  + 0.5L_{\text{GAN, rec}}(Y_{ref}, Y_{SAE}))
\end{split}
\end{equation}
\section{Results and Conclusion}
\begin{table}
\centering
\resizebox{8cm}{!}{
\begin{tabular}{|c|c c c|}
 \hline
 \textbf{Method} & FID $\downarrow$ & LPIPS $\downarrow$ & Time Taken (s) $\downarrow$\\
  \hline
  SemanticStyleGAN & - & - & 120.602\\
  \textbf{Ours} & 9.83255  & 0.1252 & \textbf{0.01143}\\
 \hline 
\end{tabular}}
\caption{Quantitative results for our method's performance versus SemanticStyleGAN \cite{Shi_2022_CVPR}. We evaluate the perceptual similarity of edits obtained (FID, LPIPS) per ROI with respect to the corresponding input and the time taken to perform one texture edit per ROI.}
\label{table: quant}
\end{table}
\begin{figure}[h]
    \centering
    \resizebox{8cm}{!}{
      \begin{tabular}{c c c c c c}
        \Huge{\textbf{Input}} &  \Huge{\textbf{Hair}} & \Huge{\textbf{Nose}} &  \Huge{\textbf{Skin}} & \Huge{\textbf{Lips + Mouth}}  &  \Huge{\textbf{Eyes}}\\
         \includegraphics[scale = 0.4]{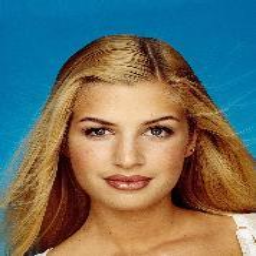} & \includegraphics[scale = 0.4]{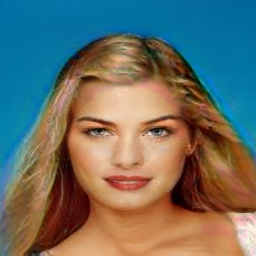} & \includegraphics[scale = 0.4]{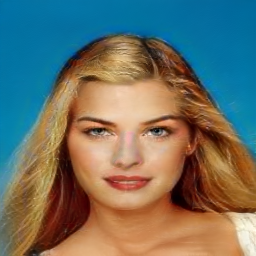} & \includegraphics[scale = 0.4]{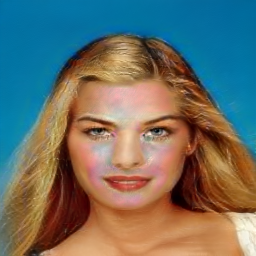} & \includegraphics[scale = 0.4]{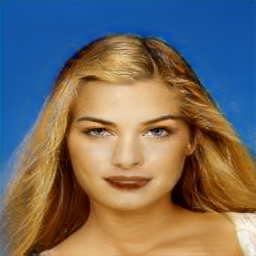} & \includegraphics[scale = 0.4]{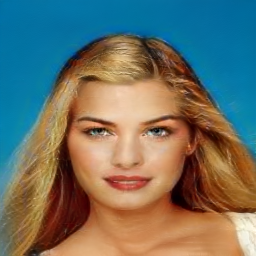} \\
         \includegraphics[scale = 0.4]{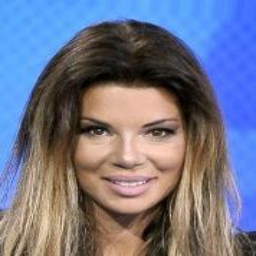} & \includegraphics[scale = 0.4]{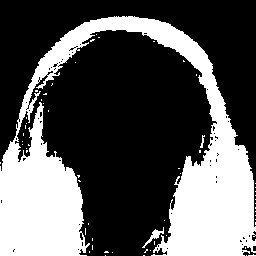} & \includegraphics[scale = 0.4]{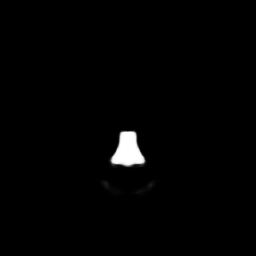} & \includegraphics[scale = 0.4]{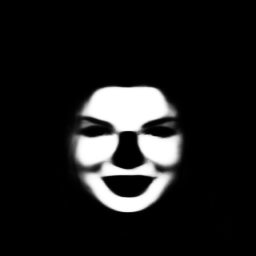} & \includegraphics[scale = 0.4]{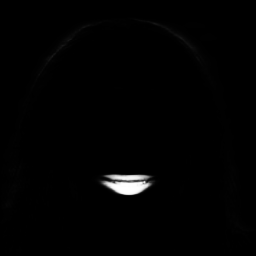} & \includegraphics[scale = 0.4]{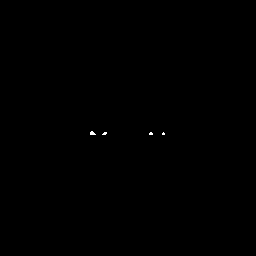} \\
\end{tabular}}
\caption{Row 1 - Localized Style Editing results. Semantic ROIs of the input image have been edited in accordance with the column label. Row 2 - SMPN ouput.}
\label{fig: results_fig}
\end{figure}
Figure \ref{fig: results_fig} and Table \ref{table: quant} present our model's qualitative and quantitative results, respectively. The perceptual metrics and qualitative results illustrate that our method performs ROI-wise localized, photorealistc, and structure preserving edits on real face images. The comparative analysis with SOTA in localized real image editing \cite{Shi_2022_CVPR} using computation time reveals the massive speed-up, our approach offers over latent-inversion based SOTA methods. Moreover, our SMPN also produces highy accurate binary semantic masks for the intended ROIs, given an input image. Dataset, implentation, and training details have been provided in the supplementary material. In conclusion,  We observe that although the results fulfill the intended objectives (clearly visible upon close inspection), the style effects are not very highly pronounced. We attribute this to the excessive refinement caused by the RB. As a future direction, we shall study ablations of the RB that yield best results and improve its architecture.
\bibliography{Formatting-Instructions-LaTeX-2023}
\end{document}